\documentclass[10pt,twocolumn,letterpaper]{article}

\usepackage{iccv}
\usepackage{times}
\usepackage{epsfig}
\usepackage{graphicx}
\usepackage{subfigure}
\usepackage{amsmath}
\usepackage{amssymb}
\usepackage{multirow}
\usepackage[british,UKenglish,USenglish,english,american]{babel}

\usepackage{algorithm}
\usepackage{algpseudocode}% http://ctan.org/pkg/algorithmicx
\algrenewcommand\algorithmicindent{.9em}%

\newcommand{\tabincell}[2]{\begin{tabular}{@{}#1@{}}#2\end{tabular}}

\usepackage[pagebackref=true,breaklinks=true,letterpaper=true,colorlinks,bookmarks=false]{hyperref}

\iccvfinalcopy % *** Uncomment this line for the final submission

\def\iccvPaperID{132} % *** Enter the ICCV Paper ID here

\ificcvfinal\fi
\begin{document}

%%%%%%%%% TITLE
\title{Flow-Guided Feature Aggregation for Video Object Detection}

\author{Xizhou Zhu$^{1,2}$\thanks{This work is done when Xizhou Zhu and Yujie Wang are interns at Microsoft Research Asia} \qquad Yujie Wang$^{2}$$^{*}$ \qquad Jifeng Dai$^{2}$ \qquad Lu Yuan$^{2}$ \qquad Yichen Wei$^{2}$ \vspace{8pt}\\
	$^{1}$University of Science and Technology of China \qquad\qquad $^{2}$Microsoft Research\qquad\qquad\\
	\hspace{0.7in}{\tt\small ezra0408@mail.ustc.edu.cn} \qquad\qquad  {\tt\small \{v-yujiwa,jifdai,luyuan,yichenw\}@microsoft.com} \qquad\\
}

%\author{Xizhou Zhu\thanks{This work is done when Xizhou Zhu and Yujie Wang are interns at Microsoft Research Asia} \qquad\qquad Yujie Wang$^{*}$ \qquad\qquad Jifeng Dai \qquad\qquad Lu Yuan \qquad\qquad Yichen Wei \vspace{8pt}\\
%	Microsoft Research Asia\\
%	{\tt\small \{v-xizzhu,v-yujiwa,jifdai,luyuan,yichenw\}@microsoft.com}
%}

\maketitle
%\thispagestyle{empty}

%%%%%%%%% ABSTRACT
\begin{abstract}  
Extending state-of-the-art object detectors from image to video is challenging. The accuracy of detection suffers from degenerated object appearances in videos, \eg, motion blur, video defocus, rare poses, etc. Existing work attempts to exploit temporal information on box level, but such methods are not trained end-to-end. We present flow-guided feature aggregation, an accurate and end-to-end learning framework for video object detection. It leverages temporal coherence on feature level instead. It improves the per-frame features by aggregation of nearby features along the motion paths, and thus improves the video recognition accuracy. Our method significantly improves upon strong single-frame baselines in ImageNet VID~\cite{russakovsky2015imagenet}, especially for more challenging fast moving objects. Our framework is principled, and on par with the best engineered systems winning the ImageNet VID challenges 2016, without additional bells-and-whistles. The proposed method, together with Deep Feature Flow~\cite{zhu2016dff}, powered the winning entry of ImageNet VID challenges 2017\footnote{\url{http://image-net.org/challenges/LSVRC/2017/results}}. The code is available at \url{https://github.com/msracver/Flow-Guided-Feature-Aggregation}.
\end{abstract}

%%%%%%%%% BODY TEXT
\section{Introduction}

Recent years have witnessed significant progress in object detection~\cite{girshick2014rich}. State-of-the-art methods share a similar two-stage structure. Deep Convolutional Neural Networks (CNNs)~\cite{krizhevsky2012imagenet,simonyan2015very,szegedy2015going,he2016deep} are firstly applied to generate a set of feature maps over the whole input image. A shallow detection-specific network~\cite{he2014spatial,girshick2015fast,ren2015faster,liu2016ssd,dai2016rfcn} then generates the detection results from the feature maps.

\begin{figure}
\begin{center}
\includegraphics[width=\linewidth]{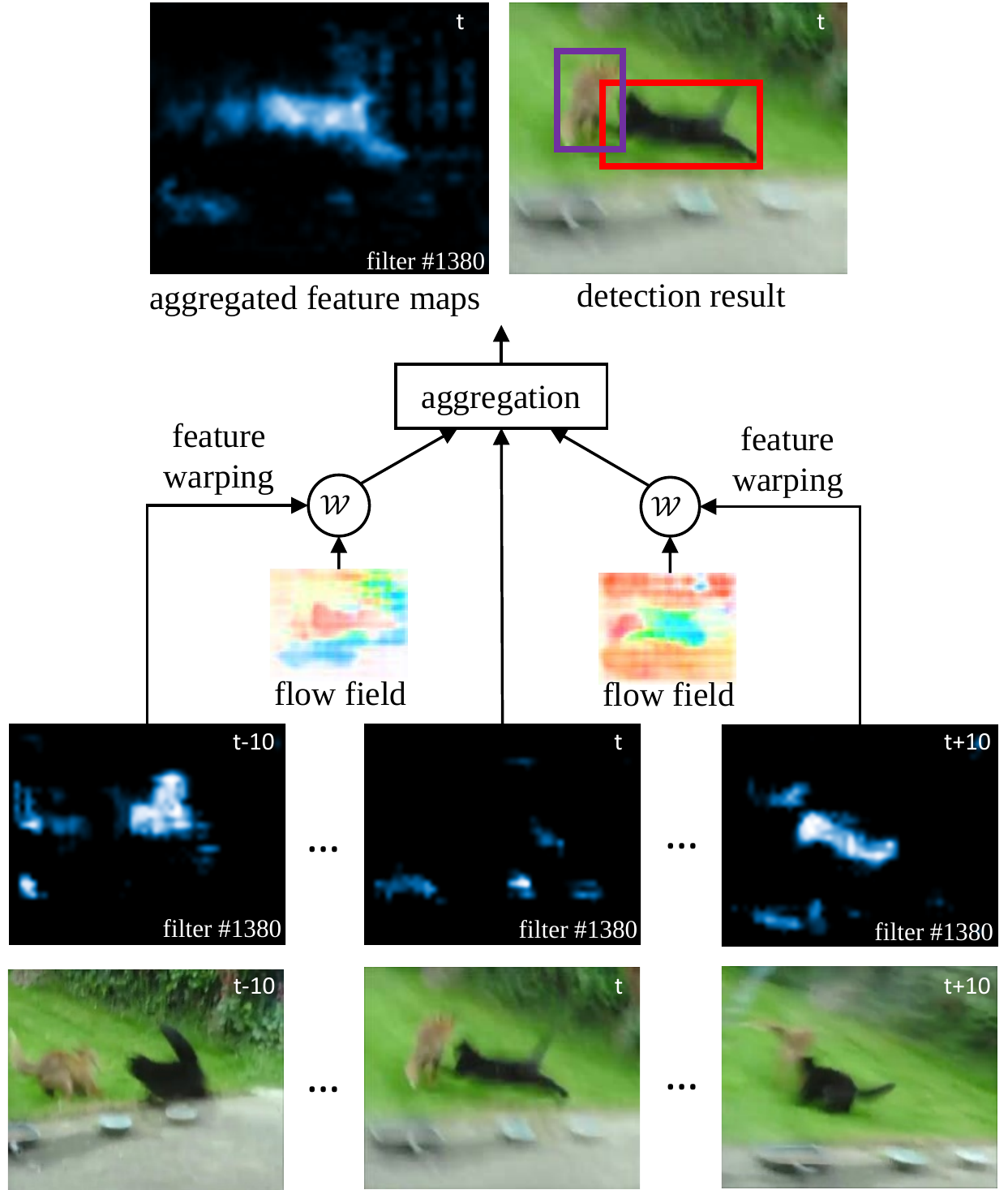}
\end{center}
\label{fig.motivation}
\caption{Illustration of FGFA (flow-guided feature aggregation). For each input frame, a feature map sensitive to ``cat'' is visualized. The feature activations are low at the reference frame $t$, resulting in detection failure in the reference frame. The nearby frames $t-10$ and $t+10$ have high activations. After FGFA, the feature map at the reference frame is improved and detection on it succeeds.} \vspace{-0.7em}
\end{figure}

\begin{figure}
\begin{center}
\includegraphics[width=\linewidth]{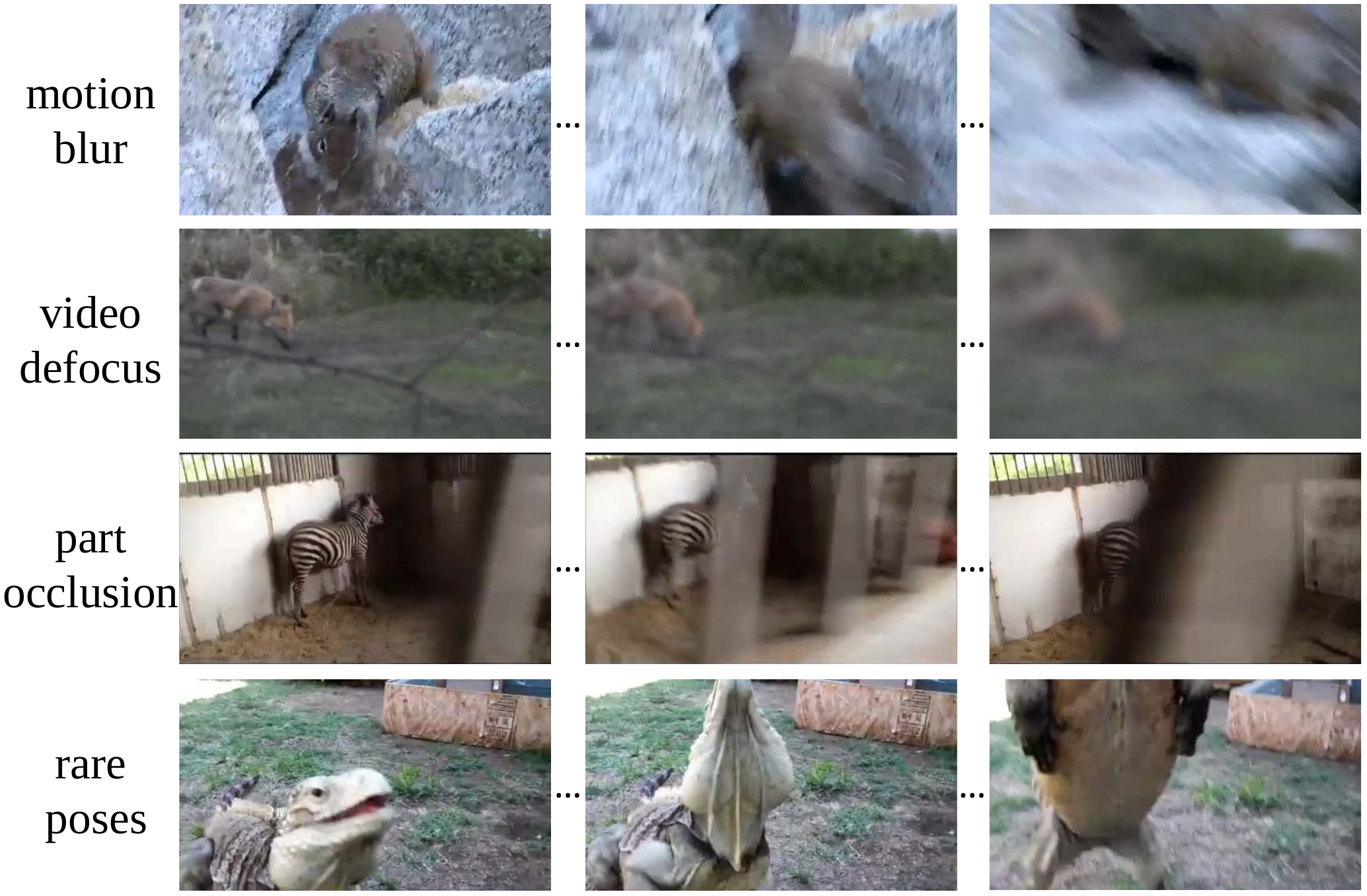}
\end{center}
\caption{Typical deteriorated object appearance in videos.}
\label{fig.challenges}
\end{figure}

These methods achieve excellent results in still images. However, directly applying them for video object detection is challenging. The recognition accuracy suffers from deteriorated object appearances in videos that are seldom observed in still images, such as motion blur, video defocus, rare poses, etc (See an example in Figure~\ref{fig.motivation} and more in Figure~\ref{fig.challenges}). As quantified in experiments, a state-of-the-art still-image object detector (R-FCN~\cite{dai2016rfcn} + ResNet-101~\cite{he2016deep}) deteriorates remarkably for fast moving objects (Table~\ref{tab.ablation_important} (a)).

Nevertheless, the video has rich information about the same object instance, usually observed in multiple ``snapshots" in a short time. Such temporal information is exploited in existing video object detection methods~\cite{kang2016tcnn,kang2016object,han2016seqnms,lee2016multi} in a simple way. These methods firstly apply object detectors in single frames and then assemble the detected bounding boxes across temporal dimension in a dedicated post processing step. This step relies on off-the-shelf motion estimation such as optical flow, and hand-crafted bounding box association rules such as object tracking. In general, such methods manipulate the single-frame detection boxes with mediocre qualities but do not improve the detection quality. The performance improvement is from heuristic post-processing instead of principled learning. There is no end-to-end training. In this work, these techniques are called \emph{box level methods}.

We attempt to take a deeper look at video object detection. We seek to improve the detection or recognition quality by exploiting temporal information, in a principled way. As motivated by the success in image recognition~\cite{girshick2014rich}, \emph{feature matters}, and we propose to improve the per-frame feature learning by temporal aggregation. Note that the features of the same object instance are usually not spatially aligned across frames due to video motion. A naive feature aggregation may even deteriorate the performance, as elaborated in Table~\ref{tab.ablation_important} (b) later. This suggests that it is critical to model the motion during learning.

In this work, we propose \emph{flow-guided feature aggregation} (FGFA). As illustrated in Figure~\ref{fig.motivation}, the feature extraction network is applied on individual frames to produce the per-frame feature maps. To enhance the features at a reference frame, an optical flow network~\cite{dosovitskiy2015flownet} estimates the motions between the nearby frames and the reference frame. The feature maps from nearby frames are warped to the reference frame according to the flow motion. The warped features maps, as well as its own feature maps on the reference frame, are aggregated according to an adaptive weighting network. The resulting aggregated feature maps are then fed to the detection network to produce the detection result on the reference frame. All the modules of feature extraction, flow estimation, feature aggregation, and detection are trained end-to-end.

Compared with box level methods, our approach works on \emph{feature level}, performs end-to-end learning and is complementary (\eg, to Seq-NMS~\cite{han2016seqnms}). It improves the per-frame features and generates high quality bounding boxes. The boxes can be further refined by box-level methods. Our approach is evaluated on the large-scale ImageNet VID dataset~\cite{russakovsky2015imagenet}. Rigorous ablation study verifies that it is effective and significantly improves upon strong single-frame baselines. Combination with box-level methods produces further improvement. We report object detection accuracy on par with the best engineered systems winning the ImageNet VID challenges, without additional bells-and-whistles (\eg, model ensembling, multi-scale training/testing, \etc).

In addition, we perform an in-depth evaluation according to the object motion magnitude. The results indicate that the fast moving objects are far more challenging than slow ones. This is also where our approach gains the most. Our method can make effective use of the rich appearance information in the varied snapshots of fast moving objects.

The proposed method, together with Deep Feature Flow~\cite{zhu2016dff}, powered the winning entry of ImageNet VID challenges 2017. The code is made publicly available at \url{https://github.com/msracver/Flow-Guided-Feature-Aggregation}.

\section{Related Work}

\textbf{Object detection from image.} State-of-the-art methods for general object detection~\cite{girshick2015fast,ren2015faster,liu2016ssd,dai2016rfcn} are mainly based on deep CNNs~\cite{krizhevsky2012imagenet,simonyan2015very,szegedy2015going,he2016deep}.
In~\cite{girshick2014rich}, a multi-stage pipeline called Regions with Convolutional Neural Networks (R-CNN) is proposed for training deep CNN to classify region proposals for object detection. To speedup, ROI pooling is introduced to the feature maps shared on the whole image in SPP-Net~\cite{he2014spatial} and Fast R-CNN~\cite{girshick2015fast}. In Faster R-CNN~\cite{ren2015faster}, the region proposals are generated by the Region Proposal Network (RPN), and features are shared between RPN and Fast R-CNN. Most recently, R-FCN~\cite{dai2016rfcn} replaces ROI pooling operation on the intermediate feature maps with position-sensitivity ROI pooling operation on the final score maps, pushing the feature sharing to an extreme.

In contrast to these methods of still-image object detection, our method focuses on object detection in videos. It incorporates temporal information to improve the quality of convolutional feature maps, and can easily benefit from the improvement of still-image object detectors.

\textbf{Object detection in video.} Recently, ImageNet introduces a new challenge for object detection from videos (VID), which brings object detection into the video domain. In this challenge, nearly all of existing methods incorporate temporal information only on the final stage `` bounding-box post-processing''. T-CNN~\cite{kang2016tcnn, kang2016object} propagates predicted bounding boxes to neighboring frames according to pre-computed optical flows, and then generates tubelets by applying tracking algorithms from high-confidence bounding boxes. Boxes along the tubelets are re-scored based on tubelets classification. Seq-NMS \cite{han2016seqnms} constructs sequences along nearby high-confidence bounding boxes from consecutive frames. Boxes of the sequence are re-scored to the average confidence, other boxes close to this sequence are suppressed. MCMOT~\cite{lee2016multi} formulates the post-processing as a multi-object tracking problem. A series of hand-craft rules (\eg, detector confidences, color/motion clues, changing point detection and forward-backward validation) are used to determine whether bounding boxes belong to the tracked objects, and to further refine the tracking results. Unfortunately, all of these methods are multi-stage pipeline, where results in each stage would rely on the results from previous stages. Thus, it is difficult to correct errors produced by previous stages.

By contrast, our method considers temporal information at the feature level instead of the final box level. The entire system is end-to-end trained for the task of video object detection. Besides, our method can further incorporate such bounding-box post-processing techniques to improve the recognition accuracy.

\textbf{Motion estimation by flow.} Temporal information in videos requires correspondences in raw pixels or features to build the relationship between consecutive frames. Optical flow is widely used in many video analysis and processing. Traditional methods are dominated by variational approaches~\cite{brox2004high,horn1981determining}, which mainly address small displacements~\cite{weickert2006survey}. The recent focus is on large displacements~\cite{brox2011large}, and combinatorial matching (\eg, Deep-Flow~\cite{weinzaepfel2013deepflow}, EpicFlow~\cite{revaud2015epicflow}) has been integrated into the variational
approach. These approaches are all hand-crafted. Deep learning based methods (\eg, FlowNet~\cite{dosovitskiy2015flownet} and its successors~\cite{ranjan2016optical,ilg2016flownet2}) have been exploited for optical flow recently. The most related work to ours is deep feature flow~\cite{zhu2016dff}, which shows the information redundancy in video can be exploited to speed up video recognition at minor accuracy drop. It shows the possibility of joint training the flow sub-network and the recognition sub-network.

In this work, we focus on another aspect of associating and assembling the rich appearance information in consecutive frames to improve the feature representation, and then the video recognition accuracy. We follow the design of deep feature flow to enable feature warping across frames.

\textbf{Feature aggregation.} Feature aggregation is widely used in action recognition~\cite{sharma2015action,amlan2016adascan,li2016videolstm,yue2015beyond,sun2015human,ballas2015delving,karpathy2014large,tran2015learning} and video description~\cite{donahue2015long,yao2015describing}.
On one hand, most of these work~\cite{sharma2015action,li2016videolstm,yue2015beyond,donahue2015long,yao2015describing,ballas2015delving,fayyaz2016stfcn,siam2016convolutional} use recurrent neural network (RNNs) to aggregate features from consecutive frames. On the other hand, exhaustive spatial-temporal convolution is used to directly extract spatial-temporal features~\cite{sun2015human,karpathy2014large,tran2015learning,tran2016deep}. However, the convolutional kernel size in these methods may limit the modeling of fast-moving objects. To address this issue, a large kernel size should be considered, but it will greatly increase the parameter number, brining issues of overfitting, computational overhead and memory consumption. By contrast, our approach relies on flow-guided aggregation, and can be scalable to different types of object motion.

\textbf{Visual tracking.}  Recently, deep CNNs have been used for object tracking~\cite{Wang2015tracking,Nam2015tracking} and achieved impressive tracking accuracy. When tracking a new target, a new network is created by combining the shared layers in the pre-trained CNN with a new binary classification layer, which is online updated. Tracking is apparently different from the video object detection task, because it assumes the initial localization of an object in the first frame and it does not require predicting class labels.

\section{Flow Guided Feature Aggregation}

\subsection{A Baseline and Motivation}

Given the input video frames $\{I_i\}, i=1, \ldots, \infty$, we aim to output object bounding boxes on all the frames, $\{y_i\}, i=1, \ldots, \infty$. A baseline approach is to apply an off-the-shelf object detector to each frame individually.

Modern CNN-based object detectors share a similar structure~\cite{girshick2014rich,girshick2015fast,ren2015faster,liu2016ssd,dai2016rfcn}. A deep convolutional sub-network $\mathcal{N}_{\rm feat}$, is applied on the input image $I$, to produce feature maps $f=\mathcal{N}_{\rm feat}(I)$ on the whole image. A shallow detection-specific sub-network, $\mathcal{N}_{\rm det}$, is applied on the feature maps to generate the output, $y = \mathcal{N}_{\rm det} (f)$.

Video frames contain drastic appearance changes of the same object instance, as exemplified in Figure~\ref{fig.challenges}. Detection on single frames generates unstable results and fails when appearance is poor. Figure~\ref{fig.motivation} presents an example. The feature responses for ``cat'' category are low at the reference frame $t$ due to motion blur. This causes single frame detection failure. Observing that the nearby frames $t-10$ and $t+10$ have high responses, their features can be propagated to the reference frame. After the features on the reference frame is enhanced, detection on it succeeds.

Two modules are necessary for such feature propagation and enhancement: 1) motion-guided spatial warping. It estimates the motion between frames and warps the feature maps accordingly. 2) feature aggregation module. It figures out how to properly fuse the features from multiple frames. Together with the feature extraction and detection networks, these are the building blocks of our approach. They are elaborated below.

\subsection{Model Design}
\textbf{Flow-guided warping.} As motivated by~\cite{zhu2016dff}, given a reference frame $I_i$ and a neighbor frame $I_j$, a flow field $\mathbf{M}_{i \rightarrow j} = \mathcal{F}(I_i, I_j)$ is estimated by a flow network $\mathcal{F}$ (\eg, FlowNet~\cite{dosovitskiy2015flownet}).

The feature maps on the neighbor frame are warped to the reference frame according to the flow. The warping function is defined as
\begin{equation}
f_{j \rightarrow i} = \mathcal{W}(f_j, \mathbf{M}_{i \rightarrow j}) = \mathcal{W}(f_j, \mathcal{F}(I_i, I_j)),
\label{eq.feature_warping}
\end{equation}
where $\mathcal{W}(\cdot)$ is the bilinear warping function applied on all the locations for each channel in the feature maps, and $f_{j \rightarrow i} $ denotes the feature maps warped from frame $j$ to frame $i$.

\textbf{Feature aggregation.} After feature warping, the reference frame accumulates multiple feature maps from nearby frames (including its own). These feature maps provide diverse information of the object instances (\eg, varied illuminations/viewpoints/poses/non-rigid deformations). For aggregation, we employ different weights at different spatial locations and let all feature channels share the same spatial weight. The 2-D weight maps for warped features $f_{j \rightarrow i}$ are denoted as $w_{j \rightarrow i}$. The aggregated features at the reference frame $\bar f_i$ is then obtained as
\begin{equation}
\bar f_i = {\sum}_{j=i-K}^{i+K} w_{j \rightarrow i} f_{j \rightarrow i},
\label{eq.feature_aggregation}
\end{equation}
where $K$ specifies the range of the neighbor frames for aggregation ($K=10$ by default). Equation~\eqref{eq.feature_aggregation} is similar to the formula of attention models~\cite{rush2015attention}, where varying weights are assigned to the features in the memory buffer.

The aggregated features $\bar f_i$ are then fed into the detection sub-network to obtain the results,
\begin{equation}
y_i = \mathcal{N}_{\rm det}(\bar f_i).
\end{equation}

Compared to the baseline and previous box level methods, our method aggregates information from multiple frames \emph{before} producing the final detection results.

\textbf{Adaptive weight.} The adaptive weight indicates the importance of all buffer frames $[I_{i-K},\dots, I_{i+K}]$ to the reference frame $I_i$ at each spatial location. Specifically, at location $p$, if the warped features $f_{j \rightarrow i}(p)$ is close to the features $f_i(p)$, it is assigned to a larger weight. Otherwise, a smaller weight is assigned. Here, we use the cosine similarity metric~\cite{luo2017cosine} to measure the similarity between the warped features and the features extracted from the reference frame. Moreover, we do not directly use the convolutional features obtained from $\mathcal{N}_{\rm feat}(I)$. Instead, we apply a tiny fully convolutional network $\mathcal{E}(\cdot)$ to features $f_i$ and $f_{j \rightarrow i}$, which projects the features to a new embedding for similarity measure and is dubbed as the embedding sub-network.

We estimate the weight by
\begin{equation}
w_{j \rightarrow i}(p)= exp(\frac{f^e_{j \rightarrow i}(p)\cdot f^e_i(p)}{|f^e_{j \rightarrow i}(p)| |f^e_i(p)|}),
%w_{j \rightarrow i}(p) \propto \exp \large(\cos(\mathcal{E}(f_{j \rightarrow i}(p)), \mathcal{E}(f_i(p)) )\large),
\end{equation}
where $f^e = \mathcal{E}(f)$ denotes embedding features for similarity measurement, and the weight $w_{j \rightarrow i}$ is normalized for every spatial location $p$ over the nearby frames, $\sum_{j = i - K}^{i + K} w_{j \rightarrow i}(p) = 1$. The estimation of weight could be viewed as the process that the cosine similarity between embedding features passes through the SoftMax operation.

\begin{algorithm}[t]
\caption{Inference algorithm of flow guided feature aggregation for video object detection.}
\small
\begin{algorithmic}[1] % [1] for line numbers
\State \textbf{input}: video frames $\{I_i\}$, aggregation range $K$
\For{$k=1$ \textbf{to} $K+1$}                            \Comment{initialize feature buffer}
\State $f_k = \mathcal{N}_{\rm feat}(I_k)$
\EndFor

\For{$i=1$ \textbf{to} $\infty$}		\Comment{reference frame}

\For{$j=\max(1, i-K)$ \textbf{to} $i+K$}		\Comment{nearby frames}
\State $f_{j \rightarrow i} = \mathcal{W}(f_j, \mathcal{F}(I_i, I_j))$			\Comment{flow-guided warp}
\State $f^e_{j \rightarrow i},f^e_i = \mathcal{E}(f_{j \rightarrow i},f_i)$	\Comment{compute embedding features}
\State $w_{j \rightarrow i}= \exp(\frac{f^e_{j \rightarrow i}\cdot f^e_i}{|f^e_{j \rightarrow i}| |f^e_i|})$	\Comment{compute aggregation weight}
\EndFor

\State $\bar f_i = \sum_{j=i-K}^{i+K} w_{j \rightarrow i} f_{j \rightarrow i}$ 	\Comment{aggregate features}
\State $y_i = \mathcal{N}_{\rm det}(\bar f_i)$		\Comment{detect on the reference frame}

\State $f_{i+K+1} = \mathcal{N}_{\rm feat}(I_{i+K+1})$		\Comment{update feature buffer}
\EndFor
\State \textbf{output}: detection results $\{y_{i}\}$
\end{algorithmic}
\label{alg.FGFA_inference}
\end{algorithm}

\subsection{Training and Inference}

\textbf{Inference.} Algorithm~\ref{alg.FGFA_inference} summarizes the inference algorithm. Given an input video of consecutive frames $\{I_i\}$ and the specified aggregation range $K$, the proposed method sequentially processes each frame with a sliding feature buffer on the nearby frames (of length $2K+1$ in general, except for the beginning and the ending $K$ frames). At initial, the feature network is applied on the beginning $K+1$ frames to initialize the feature buffer (L2-L4 in Algorithm~\ref{alg.FGFA_inference}). Then the algorithm loops over all the video frames to perform video object detection, and to update the feature buffer. For each frame $i$ as the reference, the feature maps of the nearby frames in the feature buffer are warped with respect to it, and their respective aggregation weights are calculated (L6-L10). Then the warped features are aggregated and fed to the detection network for object detection (L11-L12). Before taking the $(i+1)$-th frame as the reference, the feature maps are extracted on the $(i+K+1)$-th frame and are added to the feature buffer (L13).

As for runtime complexity, the ratio of the proposed method versus the single-frame baseline is as
\begin{equation}
r = 1+\frac{(2K+1) \cdot (\mathcal{O}(\mathcal{F}) + \mathcal{O}(\mathcal{E}) + \mathcal{O}(\mathcal{W}))}{\mathcal{O}(\mathcal{N}_{\rm feat}) + \mathcal{O}(\mathcal{N}_{\rm det})},
\end{equation}
where $\mathcal{O}(\cdot)$ measures the function complexity. Typically, the complexity of $\mathcal{N_{\rm det}}$, $\mathcal{E}$ and $\mathcal{W}$ can be ignored when they are compared with $\mathcal{N_{\rm feat}}$. The ratio is approximated as: 
$r \approx 1+\frac{(2K+1) \cdot \mathcal{O}(\mathcal{F})}{\mathcal{O}(\mathcal{N}_{\rm feat})}$. The increased computation mostly comes from $\mathcal{F}$. This is affordable, because the complexity of $\mathcal{F}$ is also much lower than that of $\mathcal{N}_{\rm feat}$ in general.

\textbf{Training.} The entire FGFA architecture is fully differentiable and can be trained end-to-end. The only thing to note is that the feature warping module is implemented by bilinear interpolation and also fully differentiable w.r.t. both of the feature maps and the flow field.

\emph{Temporal dropout.} In SGD training, the aggregation range number $K$ is limited by memory. We use a large $K$ in inference but a small $K$($=2$ by default) in training. This is no problem as the adaptive weights are properly normalized during training and inference, respectively. Note that during training, the neighbor frames are randomly sampled from a large range that is equal to the one during inference. As an analogy to dropout~\cite{srivastava2014dropout} technique, this can be considered as a \emph{temporal dropout}, by discarding random temporal frames. As evidenced in Table~\ref{tab.ablation_num_frame}, this training strategy works well.

\subsection{Network Architecture}

We introduce the incarnation of different sub-networks in our FGFA model.

\textbf{Flow network.} We use FlowNet~\cite{dosovitskiy2015flownet} (``simple" version). It is pre-trained on the Flying Chairs dataset~\cite{dosovitskiy2015flownet}. It is applied on images of half resolution and has an output stride of 4. As the feature network has an output stride of 16 (see below), the flow field is downscaled by half to match the resolution of the feature maps.

\textbf{Feature network.} We adopt the state-of-the-art ResNet (-50 and -101)~\cite{he2016deep} and Inception-Resnet~\cite{szegedy2016inception} as the feature network. The original Inception-ResNet is designed for image recognition. To resolve feature misalignment issue and make it proper for object detection, We utilize a modified version dubbed as ``Aligned-Inception-ResNet'', which is described in~\cite{dai2017deformable}. The ResNet-50, ResNet-101, and the Aligned-Inception-ResNet models are all pre-trained on ImageNet classification.

The pretrained models are crafted into feature networks in our FGFA model. We slightly modify the nature of three models for object detection. We remove the ending average pooling and the fc layer, and retain the convolution layers. To increase the feature resolution, following the practice in~\cite{chen2014semantic,dai2016rfcn}, the effective stride of the last block is changed from 32 to 16. Specially, at the beginning of the last block (``conv5" for both ResNet and Aligned-Inception-ResNet), the stride is changed from 2 to 1. To retain the receptive field size, the dilation of the convolutional layers (with kernel size $>$ 1) in the last block is set as 2. Finally, a randomly initialized $3 \times 3$ convolution is applied on top to reduce the feature dimension to 1024.

\textbf{Embedding network.} It has three layers: a $1 \times 1 \times 512$ convolution, a $3 \times 3 \times 512$ convolution, and a $1 \times 1 \times 2048$ convolution. It is randomly initialized.

\textbf{Detection network.} We use state-of-the-art R-FCN~\cite{dai2016rfcn} and follow the design in~\cite{zhu2016dff}. On top of the 1024-d feature maps, the RPN sub-network and the R-FCN sub-network are applied, which connect to the first 512-d and the last 512-d features respectively.  9 anchors (3 scales and 3 aspect ratios) are utilized in RPN, and 300 proposals are produced on each image. The position-sensitive score maps in R-FCN are of $7 \times 7$ groups.

\section{Experiments}

\subsection{Experiment Setup}

\textbf{ImageNet VID dataset~\cite{russakovsky2015imagenet}.} It is a prevalent large-scale benchmark for video object detection. Following the protocols in~\cite{kang2016tcnn,lee2016multi}, model training and evaluation are performed on the 3,862 video snippets from the training set and the 555 snippets from the validation set, respectively. The snippets are fully annotated, and are at frame rates of 25 or 30 fps in general. There are 30 object categories. They are a subset of the categories in the ImageNet DET dataset.

\begin{figure}
\begin{center}
\includegraphics[width=\linewidth]{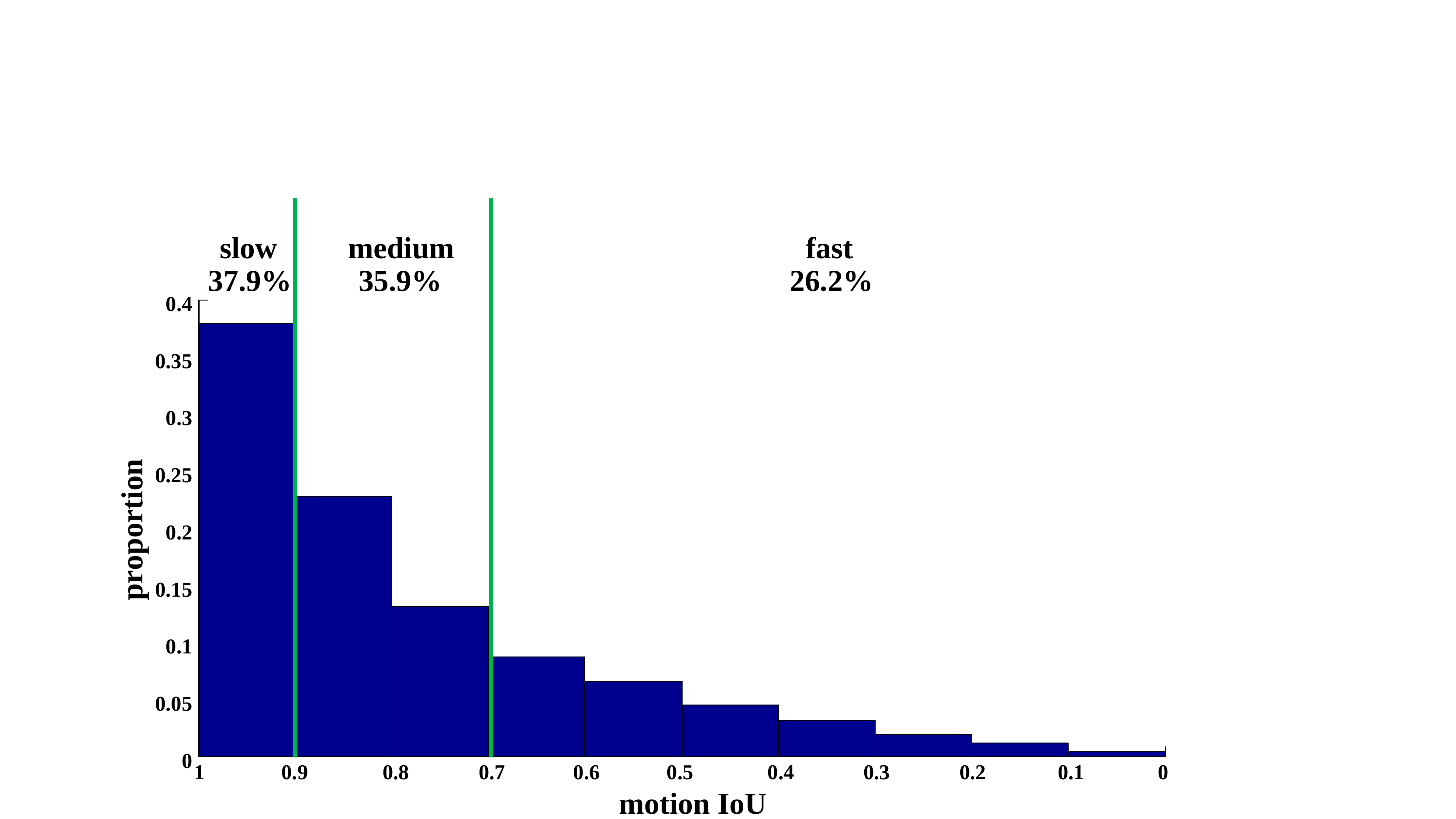}
\end{center}
\caption{Histogram of the motion IoUs of all ground truth object instances, and the division of slow, medium and fast groups.}
\label{fig.motion_distribution}
\end{figure}

\begin{figure}
\begin{center}
\includegraphics[width=\linewidth]{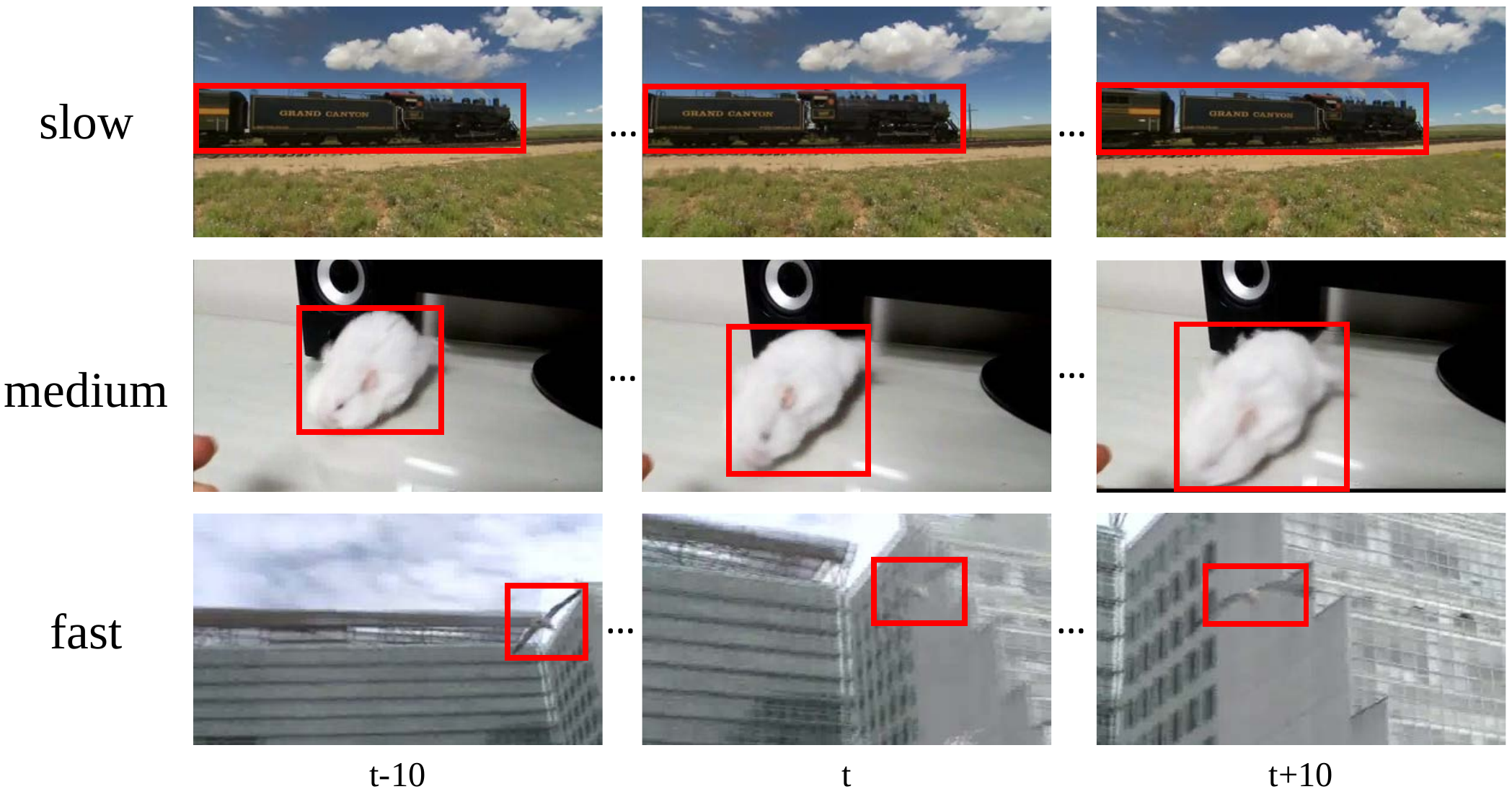}
\end{center}
\caption{Example video snippets of object instances with slow, medium and fast motions. The motion IoUs are 0.98, 0.77 and 0.26, respectively.}
\label{fig.fast_mid_slow}\vspace{-0.5em}
\end{figure}

\textbf{Slow, medium, and fast motion.} For better analysis, the ground truth objects are categorized according to their motion speed. An object's speed is measured by its averaged intersection-over-union (IoU) scores with its corresponding instances in the nearby frames ($\pm 10$ frames). The indicator is dubbed as ``motion IoU". The lower the motion IoU is, the faster the object moves. Figure~\ref{fig.motion_distribution} presents the histogram of all motion IoU scores. According to the score, the objects are divided into slow (score $>0.9$), medium (score $\in[0.7, 0.9]$), and fast (score $<0.7$) groups, respectively. Examples from various groups are shown in Figure~\ref{fig.fast_mid_slow}.

In evaluation, besides the standard mean average-precision (mAP) scores, we also report the mAP scores over the slow, medium, and fast groups, respectively, denoted as mAP(slow), mAP(medium), and mAP(fast). This provides us a more detailed analysis and in-depth understanding.

\textbf{Implementation details.} During training, following \cite{kang2016tcnn,lee2016multi}, both the ImageNet DET training and the ImageNet VID training sets are utilized. Two-phase training is performed. In the first phase, the feature and the detection networks are trained on ImageNet DET,  using the annotations of the 30 categories as in ImageNet VID. SGD training is performed, with one image at each mini-batch. 120K iterations are performed on 4 GPUs, with each GPU holding one mini-batch. The learning rates are $10^{-3}$ and $10^{-4}$ in the first 80K and in the last 40K iterations, respectively. In the second phase, the whole FGFA model is trained on ImageNet VID, where the feature and the detection networks are initialized from the weights learnt in the first phase. 60K iterations are performed on 4 GPUs, with learning rates of  $10^{-3}$ and $10^{-4}$ in the first 40K and in the last 20K iterations, respectively. In both training and inference, the images are resized to a shorter side of 600 pixels for the feature network, and a shorter side of 300 pixels for the flow network. Experiments are performed on a workstation with Intel E5-2670 v2 CPU 2.5GHz and Nvidia K40 GPU.

\setlength{\tabcolsep}{4pt}
\renewcommand{\arraystretch}{1.2}
\begin{table*}
\centering
\small
\begin{tabular}{c|c|c|c|c|c|c|c|c|c|c}
	\hline
	$\mathcal{N}_{\rm feat}$					& \multicolumn{5}{c|}{ ResNet-50} & \multicolumn{5}{c}{ ResNet-101} \\
	\hline
	methods & (a) & (b) & (c) & (d) & (e) & (a) & (b) & (c) & (d) & (e) \\
	\hline
	multi-frame feature aggregation?   &								&	\checkmark	&	\checkmark	&	\checkmark	&	\checkmark	&								&	\checkmark	&	\checkmark	&	\checkmark	&	\checkmark\\
	adaptive weights?        &								&							&	\checkmark	&	\checkmark	&	\checkmark	&								&							&	\checkmark	&	\checkmark	&	\checkmark\\
	flow-guided?                   &								&								&							& \checkmark	&	\checkmark	&								&								&							& \checkmark	&	\checkmark\\
	end-to-end training?                              &									&	\checkmark	&	\checkmark	&	\checkmark	&							&								&	\checkmark	&	\checkmark	&	\checkmark	&							\\
	\hline
	mAP (\%)                      & $70.6$ & $69.6_{\downarrow 1.0}$ & $71.8_{\uparrow 1.2}$ &  $\mathbf{74.0_{\uparrow 3.4}}$ & $72.1_{\uparrow 1.5}$ & $73.4$ & $72.0_{\downarrow 1.4}$ & $74.3_{\uparrow 0.9}$ & $\mathbf{76.3_{\uparrow 2.9}}$ & $74.5_{\uparrow 1.1}$ \\
	\hline
	mAP (\%) (slow)         & $79.3$ & $81.4_{\uparrow 2.1}$ & $81.5_{\uparrow 2.2}$ & $\mathbf{82.4_{\uparrow 3.1}}$ & $81.3_{\uparrow 2.0}$ & $82.4$ & $82.3_{\downarrow 0.1}$ & $82.2_{\downarrow 0.2}$ & $\mathbf{83.5_{\uparrow 1.2}}$ & $82.5_{\uparrow 0.1}$  \\
	\hline
	mAP (\%) (medium) & $68.6$ & $71.4_{\uparrow 2.8}$ & $71.4_{\uparrow 2.8}$ &  $\mathbf{72.6_{\uparrow 4.0}}$ & $71.5_{\uparrow 2.9}$ & $71.6$ & $74.5_{\uparrow 2.9}$ & $74.6_{\uparrow 3.0}$ & $\mathbf{75.8_{\uparrow 4.2}}$ & $74.6_{\uparrow 3.0}$ \\
	\hline
	mAP (\%) (fast)           & $50.1$ & $42.5_{\downarrow 7.6}$ & $50.4_{\uparrow 0.3}$ & $\mathbf{55.0_{\uparrow 4.9}}$ & $51.2_{\uparrow 1.1}$ & $51.4$ & $44.6_{\downarrow 6.8}$ & $52.3_{\uparrow 0.9}$ & $\mathbf{57.6_{\uparrow 6.2}}$ & $53.2_{\uparrow 1.8}$ \\
	\hline
	runtime (ms) & 203 & 204 & 220 & 647 & 647 & 288 & 288 & 305 & 733 & 733\\
	\hline 
\end{tabular}
\caption{Accuracy and runtime of different methods on ImageNet VID validation, using ResNet-50 and ResNet-101 feature extraction networks. The relative gains compared to the single-frame baseline (a) are listed in the subscript. }
\label{tab.ablation_important}
\end{table*}

\setlength{\tabcolsep}{8pt}
\renewcommand{\arraystretch}{1.2}
\begin{table}
\centering
\small
\begin{tabular}{c|c|c|c}
\hline
instance size & small & middle & large\\
\hline \hline
mAP (\%)  & 24.2 & 49.5 & 83.2 \\
\hline
mAP (\%) (slow) & 36.7 & 56.4 & 86.9\\
\hline
mAP (\%) (medium) & 32.4 & 51.4 & 80.9\\
\hline
mAP (\%) (fast)  & 24.9 & 43.7 & 67.5\\
\hline
\end{tabular}
\caption{Detection accuracy of small (area$< 50^2$ pixels), medium ($50^2 \le$area$\le 150^2$pixels), and large (area$>150^2$pixels) object instances of the single-frame baseline (entry (a)) in Table~\ref{tab.ablation_important}.}
\label{tab.ablation_size}
\end{table}

\setlength{\tabcolsep}{4pt}
\renewcommand{\arraystretch}{1.2}
\begin{table*}
\centering
\small
\begin{tabular}{c|c|c|c|c|c|c|c|c|c|c|c|c|c|c}
\hline
 \# training frames & \multicolumn{7}{c|}{ 2*} & \multicolumn{7}{c}{ 5} \\
\hline
\# testing frames & 1 & 5 & 9 & 13 & 17 & 21* & 25 & 1 & 5 & 9 & 13 & 17 & 21 & 25 \\
\hline \hline
mAP (\%) & 70.6 & 72.3 & 72.8 & 73.4 & 73.7 & 74.0 & 74.1 & 70.6 & 72.4 & 72.9 & 73.3 & 73.6 & 74.1 & 74.1\\
\hline
runtime (ms) & 203 & 330 & 406 & 488 & 571 & 647 & 726 & 203 & 330 & 406 & 488 & 571 & 647 & 726 \\
\hline
\end{tabular}
\caption{Results of using different number of frames in training and inference, using ResNet-50. Default parameters are indicated by *.}
\label{tab.ablation_num_frame}
\end{table*}

\begin{figure}
\begin{center}
\includegraphics[width=0.49\linewidth]{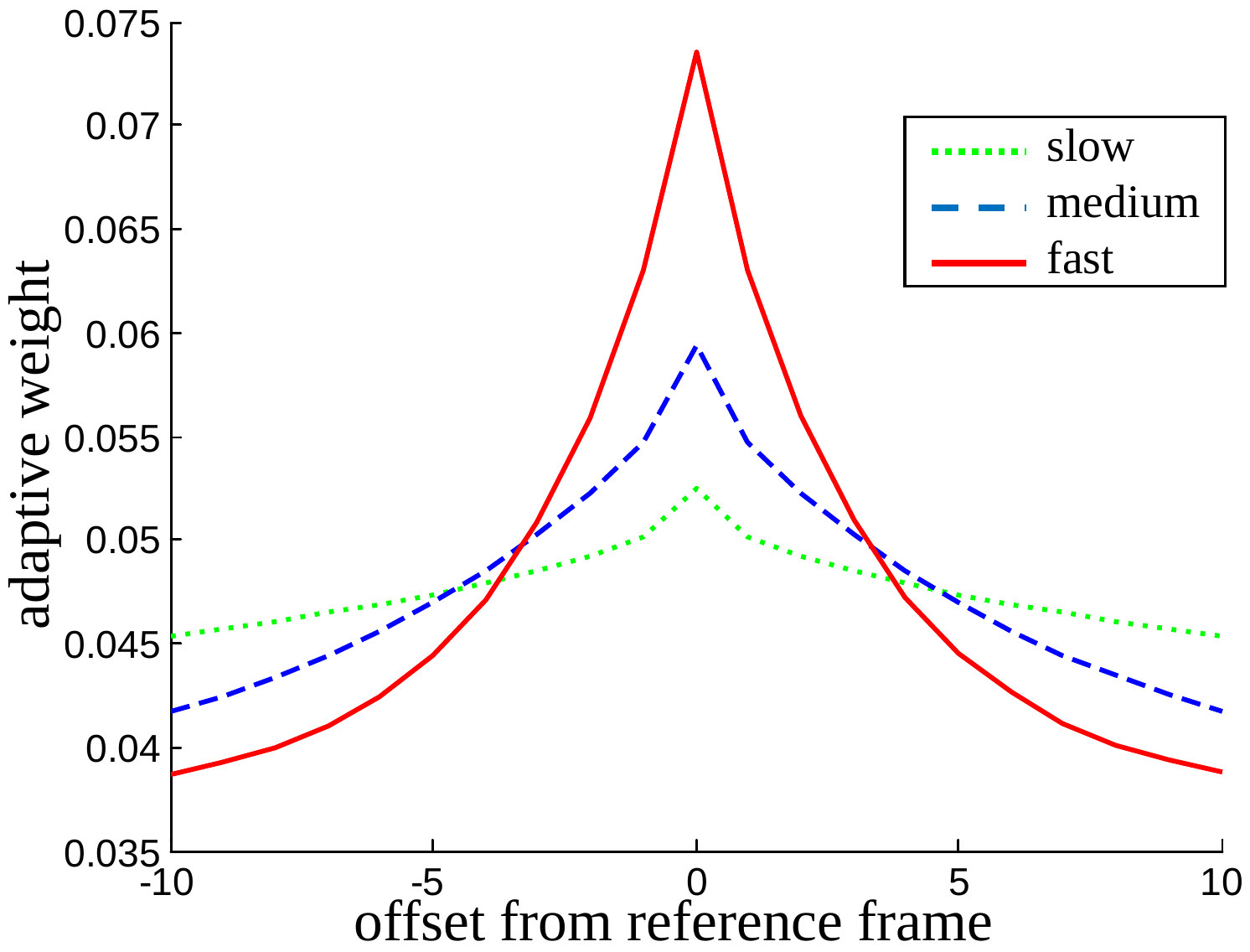} \includegraphics[width=0.49\linewidth]{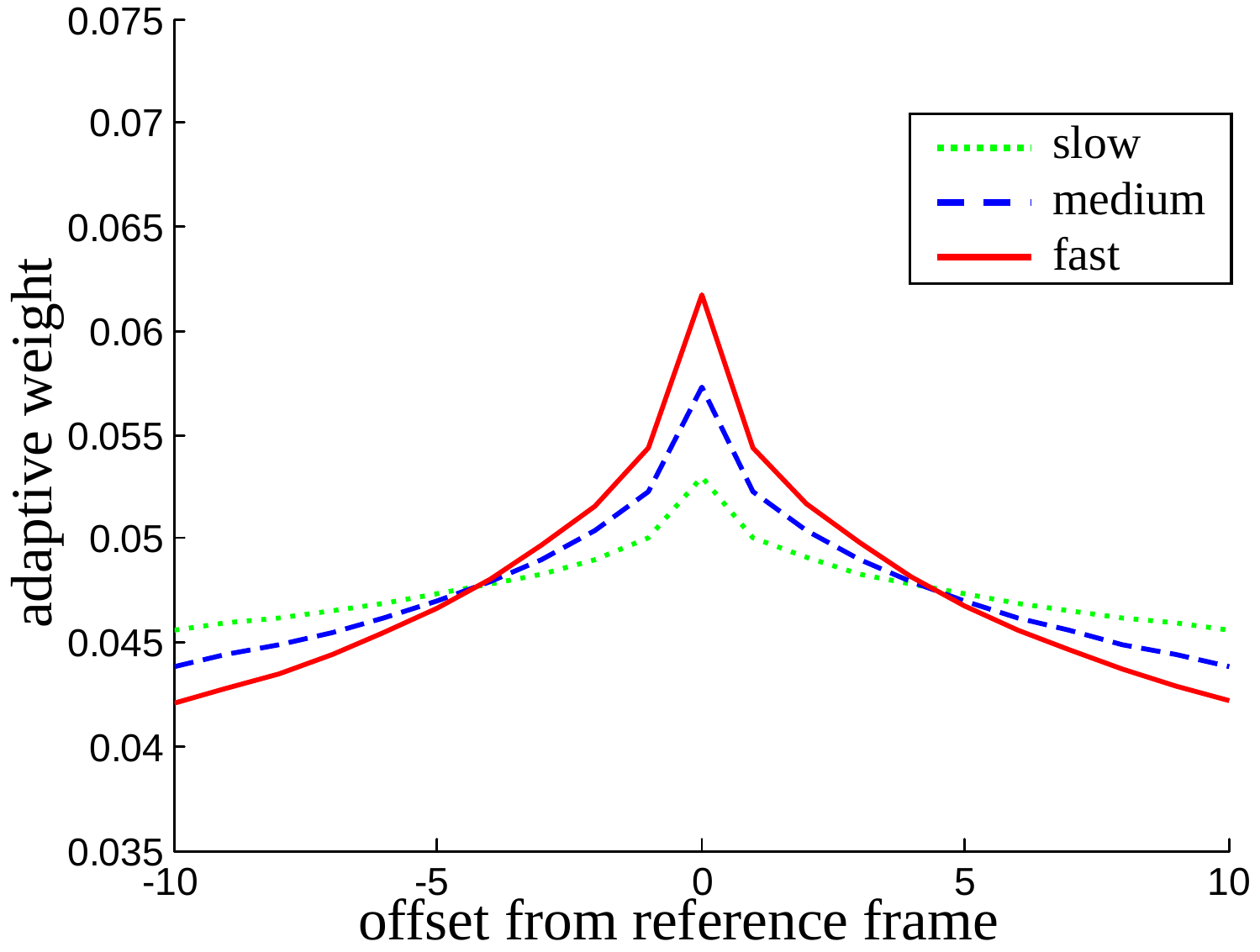} 
\end{center}
\caption{Adaptive weight distribution over frames. Left: entry without flow-guided feature warping (Table~\ref{tab.ablation_important} (c)); Right: entry with flow-guided feature warping (Table~\ref{tab.ablation_important} (d)). The histogram is performed within the boxes of instances with varying motions.}
\label{fig.adaptive_weights}\vspace{-0.5em}
\end{figure}

\subsection{Ablation Study}

\textbf{FGFA Architecture Design} Table~\ref{tab.ablation_important} compares our FGFA with the single-frame baseline and its variants.

\emph{Method (a)} is the single-frame baseline. It has a mAP 73.4\% using ResNet-101. It is close to the 73.9\% mAP in~\cite{zhu2016dff}, which is also based on R-FCN and ResNet-101. This indicates that our baseline is competitive and serves as a valid reference for evaluation. Note that we do not add bells and whistles like multi-scale training/testing, exploiting context information, model ensemble, \etc, in order to facilitate comparison and draw clear conclusions.

Evaluation on motion groups shows that detecting fast moving objects is very challenging: mAP is 82.4\% for slow motion, and it drops to 51.4\% for fast motion. As objects of different sizes may have different motion speed, we further analyze the influence of the object size. Table~\ref{tab.ablation_size} presents the mAP scores of small, middle, and large objects of different motion speeds. It shows that ``fast motion'' is an intrinsic challenge, irrespective to how large the object is.

\emph{Method (b)} is a naive feature aggregation approach and a degenerated variant of FGFA. No flow motion is used. The flow map $M_{i \rightarrow j}$ is set to all zeros in Eq.~\eqref{eq.feature_warping}. No adaptive weighting is used. The weight $w_{i \rightarrow j}$ is set to $\frac{1}{2K+1}$ in Eq.~\eqref{eq.feature_aggregation}. The variant is also trained end-to-end in the same way as FGFA. The mAP decreases to 72.0\% using ResNet-101, 1.4\% shy of baseline (a). The decrease for fast motion ($51.4\% \rightarrow 44.6\%$) is much more significant than that for slow motion ($82.4\% \rightarrow 82.3\%$). It indicates that it is critical to consider motion in video object detection.

\emph{Method (c)} adds the adaptive weighting module into (b). It obtains a mAP 74.3\%, 2.3\% higher than that of (b). Note that adding the adaptive weighting scheme is of little help for mAP (slow) and mAP (medium), but is important for mAP (fast) ($44.6\% \rightarrow 52.3\%$). Figure~\ref{fig.adaptive_weights} (Left) shows that the adaptive weights for the fast moving instances  concentrate on the frames close to the reference, which have relatively small displacement w.r.t. the reference in general.

\begin{figure*}[t]
\begin{center}
\includegraphics[width=0.9\linewidth]{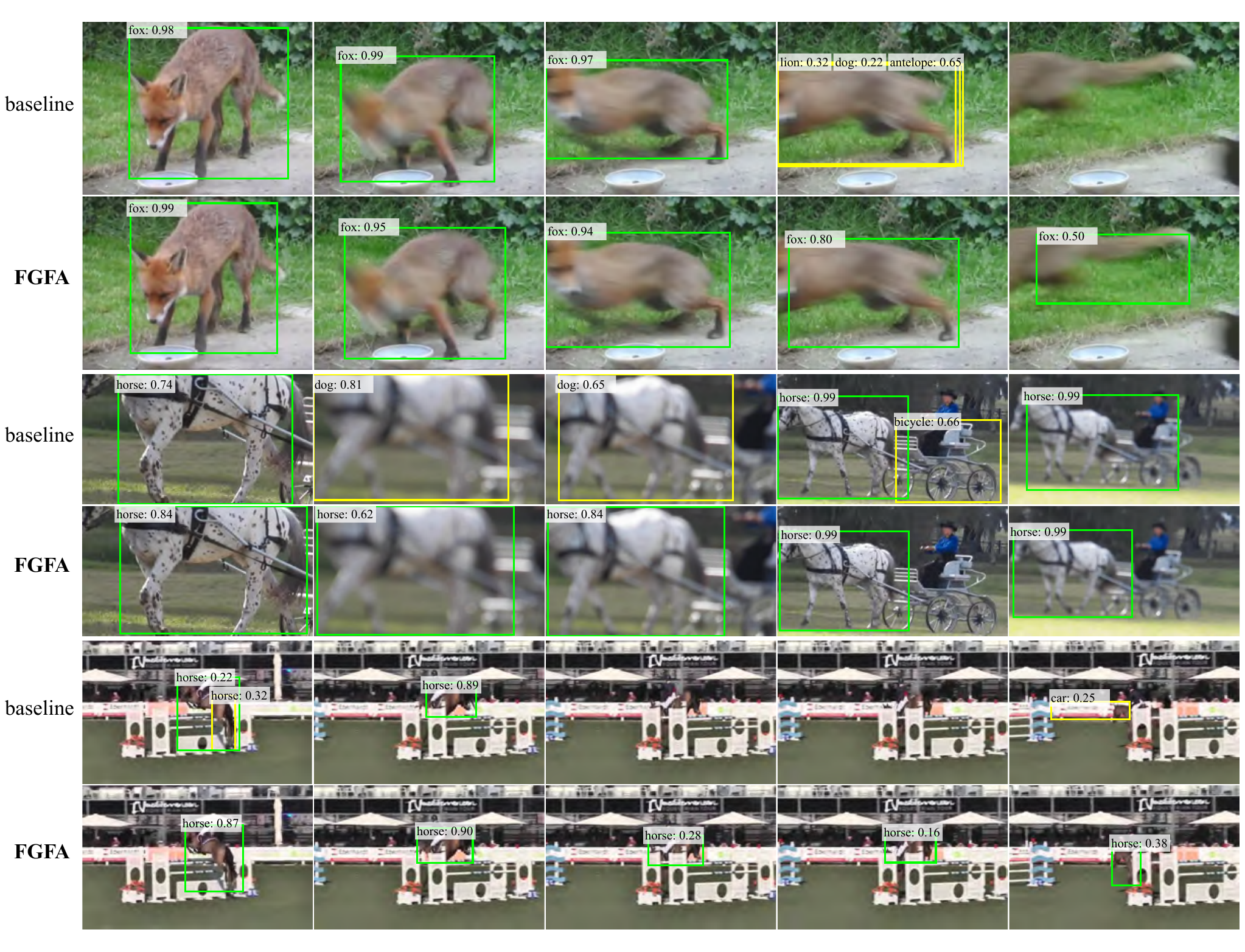}
\end{center}
\caption{Example video clips where the proposed FGFA method improves over the single-frame baseline (using ResNet-101). The green and yellow boxes denote correct and incorrect detections, respectively. More examples are available at \url{https://youtu.be/R2h3DbTPvVg}.}
\label{fig.detection_results}
\end{figure*}

\emph{Method (d)} is the proposed FGFA method, which adds the flow-guided feature aggregation module to (c). It increases the mAP score by 2\% to 76.3\%. The improvement for fast motion is more significant ($52.3\% \rightarrow 57.6\%$). Figure~\ref{fig.adaptive_weights} shows that the adaptive weights in (d) distribute more evenly over neighbor frames than (c), and it is most noticable for fast motion. It suggests that the flow-guided feature aggregation effectively promotes the information from nearby frames in feature aggregation. The proposed FGFA method improves the overall mAP score by  2.9\%, and mAP (fast) by 6.2\% compared to the single-frame baseline (a). Some example results are shown in Figure~\ref{fig.detection_results}.

\emph{Method (e)} is a degenerated version of (d) without using end-to-end training. It takes the feature and the detection sub-networks from the single-frame baseline (a), and the pre-trained off-the-shelf FlowNet. During training, these modules are fixed and only the embedding sub-network is learnt. It is clearly worse than (d). This indicates the importance of end-to-end training in FGFA.

As to runtime, the proposed FGFA method takes 733ms to process one frame, using ResNet-101 and FlowNet. It is slower than the single-frame baseline (288ms) because the flow network is evaluated $2K+1 (K = 10)$ times for each frame. To reduce the number of evaluation, we also experimented with another version of FGFA, in which the flow network is only applied on adjacent frame pairs. The flow field between non-adjacent frames is obtained by compositing the intermediate flow fields. In this way, the flow field computation on each adjacent frame pair can be re-used for different reference frames. The per-frame computation time of FGFA is reduced to 356ms, much faster than 733ms. The accuracy is slightly decreased ($\sim 1\%$) due to error accumulation in flow field composition.

\textbf{\# frames in training and inference} Due to memory issues, we use the lightweight ResNet-50 in this experiment. We tried 2 and 5 frames in each mini-batch during SGD training (5 frame reaches the memory cap), and 1, 5, 9, 13, 17, 21, and 25 frames in inference. Results in Table~\ref{tab.ablation_num_frame} show that training with 2 and 5 frames achieves very close accuracy. This verifies the effectiveness of our \emph{temporal dropout} training strategy. In inference, as expected, the accuracy improves as more frames are used. The improvement saturates at 21 frames. By default, we sample 2 frames in training and aggregate over 21 frames in inference.

\subsection{Combination with Box-level Techniques}

Our approach focuses on improving feature quality and recognition accuracy in video frames. The output object boxes can be further improved by previous box-level techniques as post-processing. In particular, we tested three prevalent techniques, namely, motion guided propagation (MGP)~\cite{kang2016tcnn}, Tubelet rescoring~\cite{kang2016tcnn}, and Seq-NMS~\cite{han2016seqnms}. Note that MGP and Tubelet rescoring are used in the winning entry of ImageNet VID challenge 2015~\cite{kang2016tcnn}. We utilized the official public code for MGP and Tubelet rescoring, and re-implemented Seq-NMS.

Table~\ref{tab.comparison_sota} presents the results. The three techniques are firstly combined with our single-frame baseline using ResNet-101 model. They all improve the baseline. This indicates that such post-processing techniques are effective. Between them, Seq-NMS obtains the largest gain. When they are combined with FGFA using ResNet-101 model, no improvement is observed for MGP and Tubelet rescoring. However, Seq-NMS is still effective (mAP increased to 78.4\%). By using  Aligned-Inception-ResNet as the feature network, the mAP of FGFA+Seq-NMS is further improved to 80.1\%, showing that Seq-NMS is highly complementary to FGFA.

\textbf{Comparison with state-of-the-art systems} Unlike image object detection, the area of video object detection lacks principled metrics~\cite{zhang2016stability} and guidelines for evaluation and comparison. Existing leading entries in ImageNet VID challenge 2015 and 2016 show impressive results, but they are complex and highly engineered systems with various bells and whistles. This makes direct and fair comparison between different works difficult.

This work aims at a principled learning framework for video object detection instead of the best system. The solid improvement of FGFA over a strong single frame baseline verifies the effectiveness of our approach. As a reference, the winning entry of ImageNet VID challenge 2016 (NUIST Team)~\cite{deng2016nuist} obtains 81.2\% mAP on ImageNet VID validation. It uses various techniques like model ensembling, cascaded detection, context information, and multi-scale inference. In contrast, our approach does not use these techniques (only Seq-NMS is used) and achieves best mAP at 80.1\%. Thus, we conclude that our approach is highly competitive with even the currently best engineered system.

\setlength{\tabcolsep}{2pt}
\renewcommand{\arraystretch}{1.2}
\begin{table}
	\centering
	\small
	\begin{tabular}{l|c|c|c}
		\hline
		method &  feature network	& mAP (\%)  & runtime (ms) \\
		\hline\hline
		single-frame baseline & \multirow{4}{*}{ResNet-101} & 73.4 & 288 \\
		+ MGP & & 74.1 & 574* \\
		+ Tubelet rescoring & & 75.1 & 1662 \\
		+ Seq-NMS & & 76.8 & 433* \\
		\hline
		\textbf{FGFA} & \multirow{4}{*}{ResNet-101} & 76.3 & 733 \\
		+ MGP & & 75.5 & 1019* \\
		+ Tubelet rescoring & & 76.6 & 1891 \\
		+ Seq-NMS & & \underline{78.4} & 873* \\
		\hline
		\textbf{FGFA} & \multirow{2}{*}{\tabincell{c}{Aligned- \\ Inception-ResNet}} & 77.8 & 819 \\
		+ Seq-NMS & & \textbf{80.1} & 954* \\
		\hline 
	\end{tabular}
	\caption{Results of baseline method and FGFA before and after combination with box level techniques. As for runtime, entry marked with * utilizes CPU implementation of box-level techniques.}
	\label{tab.comparison_sota}
\end{table}

\section{Conclusion and Future Work}

This work presents an accurate, end-to-end and principled learning framework for video object detection. Because our approach focuses on improving feature quality, it would be complementary to existing box-level framework for better accuracy in video frames. Several important aspects are left for further exploration. Our method slows down a bit, and it would be possibly sped up by more lightweight flow networks. There is still large room to be improved in fast object motion. More annotation data (\eg, YouTube-BoundingBoxes~\cite{Esteban2017youtube}) and precise flow estimation may be benefit to improvements. Our method can further leverage better adaptive memory scheme in the aggregation instead of the attention model used. We believe these open questions will inspire more future work.

{\small
\bibliographystyle{ieee}
\bibliography{egbib}
}

\end{document}